\documentclass[runningheads]{llncs}

\usepackage{eccv}

\usepackage{eccvabbrv}

\usepackage{graphicx}
\usepackage{booktabs}

\usepackage[accsupp]{axessibility}  

\usepackage{hyperref}

\usepackage{orcidlink}

\begin{document}

\title{CC-SAM: SAM with Cross-feature Attention and Context for Ultrasound Image Segmentation} 


\author{Shreyank N Gowda\inst{1} \and
David A. Clifton\inst{1,2}}
%
\authorrunning{Shreyank N Gowda and David A. Clifton}
%
\institute{Department of Engineering Sciences, University of Oxford, OX3 7DQ Oxford, UK \and
Oxford Suzhou Centre for Advanced Research, University of Oxford, Suzhou 215123, Jiangsu, China}

\maketitle

\begin{abstract}
The Segment Anything Model (SAM) has achieved remarkable successes in the realm of natural image segmentation, but its deployment in the medical imaging sphere has encountered challenges. Specifically, the model struggles with medical images that feature low contrast, faint boundaries, intricate morphologies, and small-sized objects. To address these challenges and enhance SAM's performance in the medical domain, we introduce a comprehensive modification. Firstly, we incorporate a frozen Convolutional Neural Network (CNN) branch as an image encoder, which synergizes with SAM's original Vision Transformer (ViT) encoder through a novel variational attention fusion module. This integration bolsters the model's capability to capture local spatial information, which is often paramount in medical imagery. Moreover, to further optimize SAM for medical imaging, we introduce feature and position adapters within the ViT branch, refining the encoder's representations. We see that compared to current prompting strategies to fine-tune SAM for ultrasound medical segmentation, the use of text descriptions that serve as text prompts for SAM helps significantly improve the performance. Leveraging ChatGPT's natural language understanding capabilities, we generate prompts that offer contextual information and guidance to SAM, enabling it to better understand the nuances of ultrasound medical images and improve its segmentation accuracy. Our method, in its entirety, represents a significant stride towards making universal image segmentation models more adaptable and efficient in the medical domain.
\end{abstract}

\section{Introduction}
\label{sec:intro}

Deep learning has revolutionized the fields of image~\cite{resnet,colornet,densenet}, video classification~\cite{gowda2017human,gowda2021smart} and medical vision~\cite{unet,fatnet,zhou2018unet++,gowda2024masks}, enabling unprecedented accuracy and efficiency in these areas. Medical image segmentation plays a pivotal role in delineating and emphasizing particular organs, tissues, and anomalies within medical scans, forming a cornerstone of computer-aided diagnostic systems~\cite{liu2021review}. The development of numerous deep learning models for autonomous medical image segmentation has signaled a transformative shift in the potential of this technology~\cite{unet,fatnet,chen2018drinet,zhou2018unet++}. Nevertheless, these models are typically specialized for distinct tasks and require recalibration when used for different tasks, a limitation which poses significant challenges in medical data.

\begin{figure}[t]
  \centering
  \includegraphics[width=0.94\textwidth]{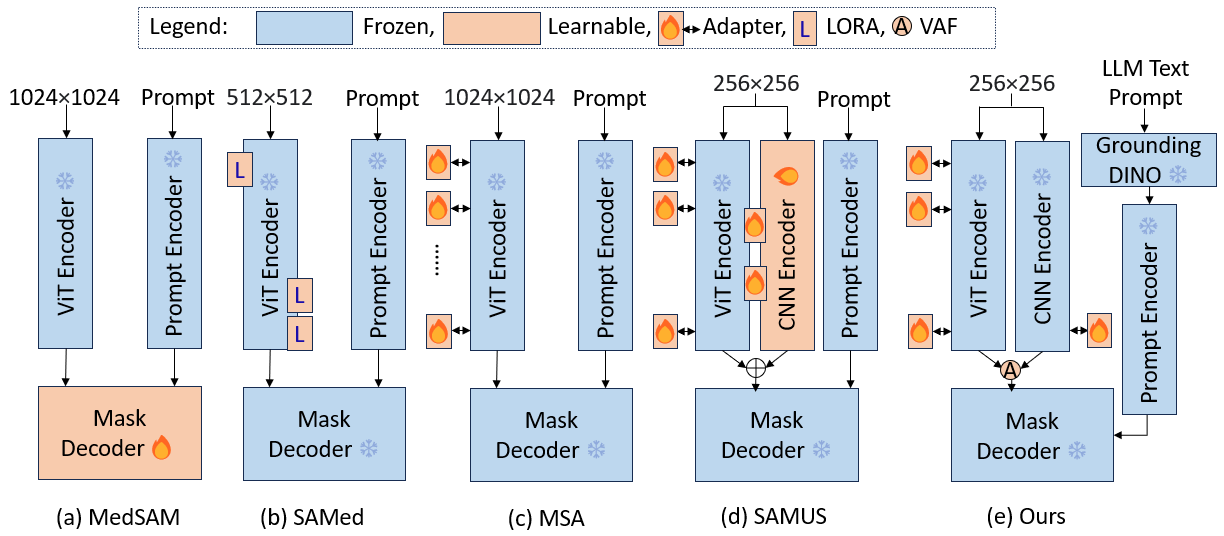}

   \caption{Comparison of methods using SAM for medical image segmentation.}
   \label{fig:teaser}
\end{figure}

Segment anything model (SAM)~\cite{sam} has emerged as a versatile foundational model for vision segmentation, gaining significant attention due to its unparalleled ability to segment a wide variety of objects and exhibit powerful zero-shot generalization. With user-driven prompts such as points, bounding boxes, and coarse masks, SAM dynamically segments the relevant objects. This adaptability positions SAM as an ideal tool for a myriad of segmentation use cases, facilitating the amalgamation of several discrete medical image segmentation tasks into a holistic model, enhancing its utility for clinical applications~\cite{huang2023segment}.

However, while SAM boasts an extensive dataset (SA1B), its performance diminishes in the medical realm due to the lack of comprehensive clinical annotations~\cite{huang2023segment}. To enhance SAM's performance in medical image segmentation, many approaches have been proposed such as tuning the mask decoder~\cite{medsam}, applying the LoRA strategy to the image encoder~\cite{samed}, introducing task-specific information using adapters in the ViT image encoder~\cite{msa}. Previous endeavors to adapt SAM for medical image segmentation by optimizing it on medical datasets have encountered limitations, given their tokenization approach that can obfuscate essential local data.

SAMUS~\cite{samus} was subsequently introduced, aiming to transplant SAM's superior segmentation prowess and robust generalization to medical image segmentation, all while ensuring computational efficiency. Notable improvements within SAMUS encompass a redesigned ViT image encoder, integration of a parallel CNN-branch, and the introduction of a cross-branch attention mechanism, among others. 

In this paper, we introduce CC-SAM, a novel model built upon SAMUS. Notably, CC-SAM replaces the adjustable CNN with a fixed one, enhanced with adapters and employs a novel variational attention fusion instead of the cross-attention branch. Utilizing descriptions from Chat-GPT to prompt the model, we achieve a marked improvement in performance over existing prompting strategies tailored for medical-oriented SAM. This new approach significantly improves medical-oriented SAM performance. See Figure \ref{fig:teaser} for a comparison with other medical SAM adaptations.

The chief contributions of this paper are delineated as follows:

\begin{itemize}
    \item The development of CC-SAM: a refined foundation model tailored for universal medical image segmentation, optimized for computational efficiency.
    \item The introduction of a static CNN with adapters, complementing the ViT encoder of SAMUS and further reducing computational costs.
    \item The innovative integration of variational attention fusion, enhancing the synergy between CNN and ViT branch features.
    \item The use of text-based prompts, sourced from Chat-GPT, to articulate the segmentation problem, which significantly amplifies segmentation performance in medical contexts.
\end{itemize}

\section{Related Work}
\label{sec:related}

\subsection{Medical Image Segmentation}


Medical image segmentation is crucial for identifying and measuring structures in medical images. Before deep learning, methods like thresholding, clustering, and active contour models were common~\cite{pham2000current}. However, CNNs, particularly U-Net~\cite{unet}, have since dominated due to their quality results even with limited data.


Extensions of U-Net, such as DRINet~\cite{chen2018drinet}, tackled class imbalance and high-resolution outputs. U-Net++\cite{zhou2018unet++} used nested and dense skip pathways to improve segmentation, while UNeXt\cite{unext} optimized for speed and efficiency in medical segmentation by using a Convolutional MLP-based network.


The rise of image transformers~\cite{vit} led to new methods. FAT-Net~\cite{fatnet} introduced feature adaptive transformers for detailed segmentation. UNeTR~\cite{unetr} combined transformers with a U-shaped design for 3D medical imaging. Our approach merges CNNs and transformers to harness both local and global feature extraction, creating a hybrid framework that leverages the strengths of both paradigms.

\subsection{Adapting Foundational Models}

Foundational models, often pre-trained on large datasets, provide a generic understanding of complex patterns in data. These models are crucial stepping stones in many modern machine learning workflows. However, leveraging them effectively for specific tasks necessitates a thoughtful adaptation strategy. 


Adapter modules~\cite{adapter} offer a solution, allowing the main network weights to remain unchanged while adding learnable layers to each Transformer layer. This adds task-specific abilities with less overhead, ensuring efficient transfer learning. This approach keeps the base knowledge while adapting to new tasks.

These modules have excelled in various language tasks~\cite{adapter2,adapter3,adapter4}. For instance, they've achieved top results in question answering with 95\% fewer parameters than standard finetuning~\cite{adapter1}. In textual entailment, AdaMix~\cite{wang2022adamix} found adapters performed comparably to full finetuning but with only 3\% added parameters. They've also been effective in computer vision tasks like image~\cite{adapter5,adapter6} and video classification~\cite{adapter7,adapter8,adapter9}. Essentially, they ensure efficient transfer learning across tasks by adding minimal layers to fixed networks.


We build on the idea of adapters and freeze both our CNN and ViT backbones and only use adapters for efficient fine-tuning of both models.

\subsection{Adapting SAM for Medical Images}

The Segment Anything Model (SAM)\cite{sam} has demonstrated notable zero-shot segmentation using various input prompts. However, its performance on medical images is subpar\cite{sammed1,sammed2}.

Several strategies have been proposed to adapt SAM for medical imaging. MedSAM optimizes the mask decoder for efficient fine-tuning. SAMed~\cite{samed} freezes the image encoder and employs a low-rank finetuning strategy while adjusting the prompt encoder and mask decoder. MSA~\cite{msa} uses adapters in the ViT image encoder. To address transformer backbone limitations, SAMUS~\cite{samus} incorporates a CNN-based backbone for improved local feature learning.

Building on SAMUS~\cite{samus}, we optimize its framework. Instead of end-to-end CNN fine-tuning, we utilize a frozen pre-trained CNN and an adapter. We found that using text prompts, especially with Grounding DINO~\cite{gdino}, significantly enhances SAMUS's performance. This method is not only more effective but also faster. We use GPT-4~\cite{gpt4} to generate text prompts from class labels.

\begin{figure*}
    \centering
    \includegraphics[width=0.95\textwidth]{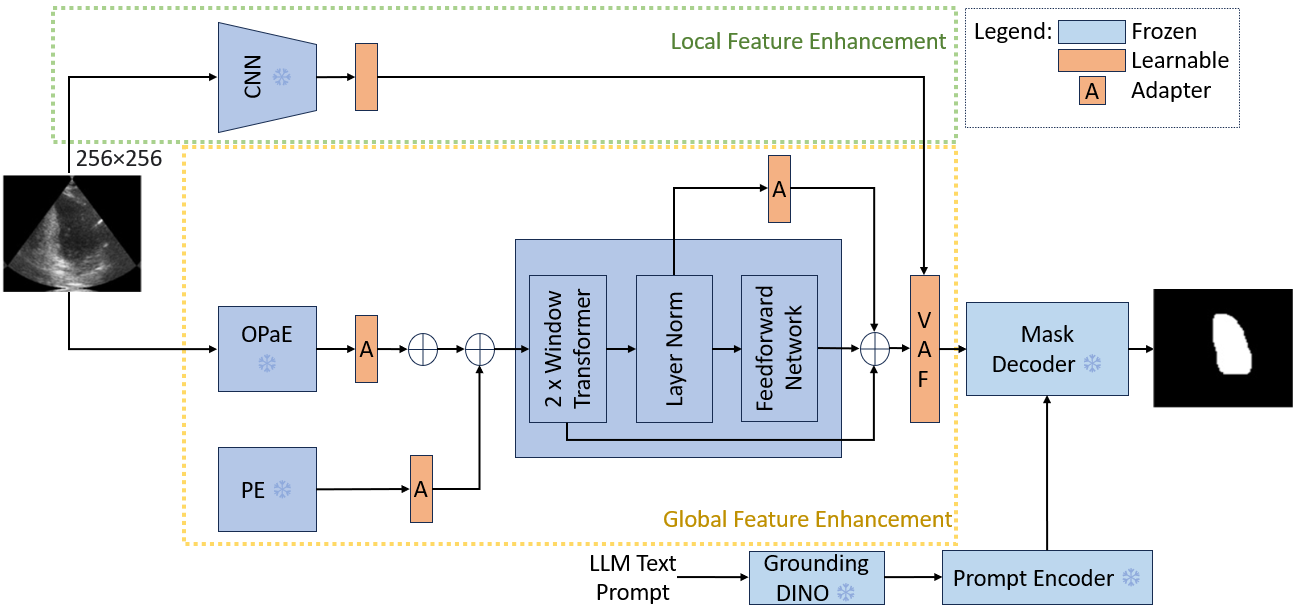}
    \caption{Overview of CC-SAM. We use adapters (or in the case of the CNN a FC layer) to enhance local and global features for ultrasound segmentation. `OPaE' refers to overlapping patch embeddings and `PE' refers to positional embeddings.}
    \label{fig:overview}
\end{figure*}

\section{Proposed Method}
\label{sec:method}



Figure~\ref{fig:overview} provides a detailed view of our method. Essentially, we process an image $I$ using ViT-B~\cite{vit} (pre-trained image encoder of SAM) and ResNet50~\cite{resnet}, the ResNet50 is pre-trained on RadImageNet~\cite{rin}. After feature extraction, adapters refine these representations. Features are then fused using a variational attention mechanism, forming a combined feature for the Mask Decoder of SAM.

For prompts, we use a GPT-4 generated label description and embed it with Med-BERT~\cite{medbert}. The Grounding DINO model, untrained on medical data, creates a bounding box from this embedding, which is input to the Prompt Encoder. Together with the variational attention feature, the Prompt Encoder helps the Mask Decoder create the final segmentation mask.

\subsection{Frozen Backbones with Adapters}



We've enhanced SAM's image encoder (the ViT branch) for better adaptability to smaller inputs and medical images by introducing a position adapter and five feature adapters, following SAMUS~\cite{samus}. These adapters fine-tune the ViT branch efficiently with fewer parameters.

Specifically, the position adapter modifies positional embeddings to match the embedded sequence resolution. It first downsamples these embeddings using max pooling and then refines them with a convolution operation, enabling the ViT to better manage smaller inputs. Each of the five feature adapters follows a consistent design with a downward linear projection, an activation, and an upward projection. This is mathematically expressed as:

\begin{equation}
    A(x)=G(x\mathcal{M}_{d})\mathcal{M}_{u}
\end{equation}

Here, Here, $G$ denotes the GELU activation function, while $\mathcal{M}_{d} \in \mathbb{R}^{d\times \frac{d}{4}}$ and $\mathcal{M}_{u} \in \mathbb{R}^{d\times \frac{d}{4}}$ are the projection matrices. In this context, $'d'$ represents the dimension of the feature embedding. 


Unlike SAMUS's end-to-end CNN training, we use a static, RadImageNet~\cite{rin} pre-trained ResNet-50 model. We add a trainable fully-connected layer, acting as an adapter, before its classification layer. The output features from both branches then go into our variational attention fusion block.

\subsection{Variational Attention Fusion Block}

\begin{figure}[t]
  \centering
  \includegraphics[width=0.75\textwidth]{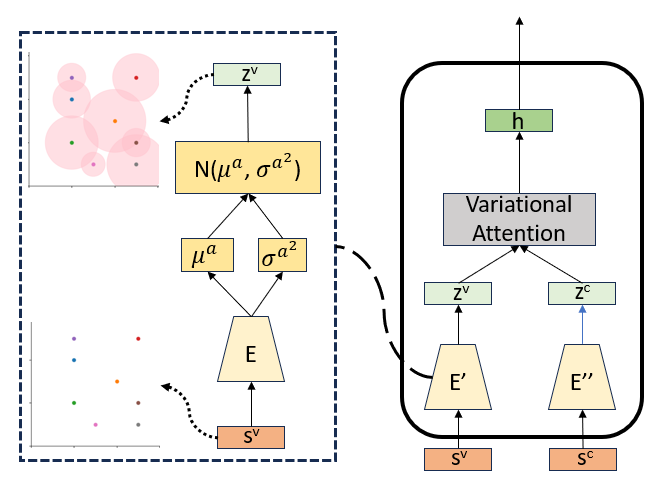}

   \caption{Overview of the proposed Variational Attention Fusion Block. Each `mode' has an intra-modal uncertainty learning encoder (represented as E' or E'' in the figure), these obtain robust
modality-specific features in the latent subspace. Subsequently, VAF combines these inputs and constructs a multimodal representation by estimating weights that are specific to each modality, effectively capturing their dependencies.}
   \label{fig:vaf}
\end{figure}


A major contribution we've made is the Variational Attention Fusion Block, which adeptly merges local CNN features and global ViT features. While SAMUS introduced a cross-branch attention module, our method, which models uncertainty around these features and applies variational attentional fusion, outperforms it (details in Sec~\ref{sec:ablation}). See Figure~\ref{fig:vaf} for an overview.

\subsubsection{Uncertainty Learning}



Incorporating uncertainty learning, for an input image $I$, we derive features $s^{v}$ from the ViT and $s^{c}$ from the ResNet-50. Both $s^{v}$ and $s^{c} \in \mathbb{R}^{d^{h}}$ represent a fixed dimension in a latent space. We achieve a consistent size using latent encoders $E_{v}, E_{c}$ at each network's end. To address data uncertainty, we design a latent distribution for each sample's content, helping capture semantic relations. This distribution is shaped as a parameterized diagonal Gaussian. This is mathematically represented as:

\begin{equation}
     p(z^{v}|s^{v})=\mathcal{N}(z^{v};\mu ^{v},(\sigma ^{v})^{2}\mathbf{I})
\end{equation}
\begin{equation}
     p(z^{c}|s^{c})=\mathcal{N}(z^{c};\mu ^{c},(\sigma ^{c})^{2}\mathbf{I})
\end{equation}



Here, $z^{v}$ and $z^{c}$ are reconstructed vectors. The means, $\mu^{v}, \mu ^{c}$, capture each mode's core feature, while variances, $\sigma^{v}, \sigma^{c}$, indicate the noise-induced uncertainty in predicting these means. A higher variance signifies more uncertainty about the content observed. Both Gaussian parameters depend on the input and are forecasted through MLPs. For instance, for ViT features: $\mu^{v} = f{\theta_{v1}}(s{v})$ and $\sigma^{v} = f{\theta_{v2}}(s{v})$, with $\theta_{v1}$ and $\theta_{v2}$ being respective model parameters. The same applies to CNN features. Now, each sample's feature representation shifts from deterministic to a stochastic Gaussian-drawn embedding in the latent space.

Due to the non-differentiable nature of sampling, we use a re-parameterization trick~\cite{repar} to maintain gradient flow. We draw random noise $\eta$ from a normal distribution, independent of model parameters, and produce $z^{v}$ using it, as detailed in Eq~\ref{eq:repar}. We repeat this for $z^{c}$.

\begin{equation}
\label{eq:repar}
    z^{v}=\mu ^{v}+\eta \sigma ^{v},\eta \in \mathcal{N}(0,1)
\end{equation}

\subsubsection{Variational Attentional Fusion}



CNNs and ViTs capture different feature aspects from images (local and global). We see these as two modes: $v$ for ViTs and $c$ for CNNs. Combining them isn't straightforward due to different confidence levels. Standard methods, which get weights from the data, miss the unique traits of each mode. To tackle this, we've developed a Variational Attention Fusion (VAF) module. This module aims to capture the nuances between modalities by determining modality-specific weights for a seamless integration.

In short, using features from each mode, a standard attention method creates a probability, $a^{k}$ (where $k$ is either $v$ or $c$), as shown in Eq~\ref{eq:attn}. Here, $a^{k}$ shows how much each mode contributes, and $\textbf{W}_{m}$ and $b_{m}$ are trainable factors. By having set weights, it picks the right features without being affected by varied confidence levels between modes.

\begin{equation}
    \label{eq:attn}
    \begin{split}
    \hat{a}^{k}=\textbf{W}_{m}\textbf{z}^{k}+b_{m} \\ a^{k}=\frac{exp(\hat{a}^{k})}{\sum_{k'\in \left\{ v,c\right\}}exp(\hat{a}^{k'})}
    \end{split}
\end{equation}


Our VAF module uses variational attention weights instead of the point-estimated attention vector $\hat{a}^{k}$. This approach, grounded in a probability distribution (Eq.~\ref{eq:vaf}), better accounts for uncertainties across modalities.

\begin{equation}
    \label{eq:vaf}
    \hat{a}^k \sim q_\theta\left(\hat{a}^k \mid z^k\right)=\mathcal{N}\left(\mu_a^k,\left(\sigma_a^k\right)^2 \mathbf{I}\right), k \in\{v, c\}
\end{equation}


We used an identity transformation to preserve modality traits in the variational attention. Mean and variance parameters are predicted by MLPs from the input. The $\left(\sigma_a^k\right)$ value indicates confidence levels among modalities. With the VAF method, more confident modality features are enhanced, while less confident ones are reduced, optimizing multimodal data fusion and capturing complementary features. The final representation combines modality-specific outputs, $z_{c}$ and $z_{v}$, using weighted aggregation as shown in Eq~\ref{eq:final}, where $\textbf{W}_{h}$ is a learnable weight matrix.

\begin{equation}
    \label{eq:final}
    h = \sum_{k \in \left\{ v,c\right\}}a^{k}\textbf{W}_{h}z^{k}
\end{equation}

\subsection{Guiding the Prompt Encoder}


SAM's mask decoder~\cite{sam} needs $h$ and an input from the prompt encoder. While past adaptations of SAM for medical imaging used point prompts, our results show that a reasonably good bounding box enhances performance. Creating this bounding box requires a dedicated model. Using random bounding boxes may hinder results, necessitating specific prompts. We employ Grounding-DINO~\cite{gdino} for bounding box generation. Though not exclusively trained for medical images, Grounding-DINO is an effective object detector. We use GPT-4~\cite{gpt4} to craft text prompts for specific class labels, and then MedBERT~\cite{medbert} generates an embedding input for Grounding-DINO. Whilst this does improve our performance in comparison to using point prompts, we show in supplementary that using the same set of random point prompts, we outperform all other medical foundational models.

\textit{{\footnotesize Input to GPT-4 "Create a concise description for a medical image analysis task. The task involves using ultrasound images to segment and identify specific anatomical structures or pathologies. Use [target] for to make a generic description that can be adapted for any segmentation target."
Output from GPT-4 "Segment and identify [target] within ultrasound images. This requires the precise delineation of [target] from surrounding tissues, enabling accurate diagnosis and assessment. The challenge encompasses dealing with the inherent variability of ultrasound image quality, including speckle noise, shadowing, and the wide range of anatomical variations among patients. Success in this task is measured by the algorithm's ability to consistently and accurately identify and outline the [target], providing crucial information for medical decision-making processes. " where [target] is modified for each dataset.}}

\subsection{Loss Function}

Our overall loss function can be seen in Eq~\ref{eq:loss}. Here, $\mathcal{L}_{BCE}$ and $\mathcal{L}_{D}$ are segmentation losses: binary cross entropy and dice loss respectively. 

\begin{equation}
    \label{eq:loss}
    \mathcal{L} = \mathcal{L}_{BCE}+\mathcal{L}_{D}+\lambda _{c}\mathcal{L}_{c}+\lambda _{v}\mathcal{L}_{v}+\lambda _{a}\mathcal{L}_{a}
\end{equation}


$\mathcal{L}{c}$ and $\mathcal{L}{v}$ are regularization terms for uncertainty learning, inspired by the variational information bottleneck~\cite{vib}. To grasp uncertainty in a modality, a component is introduced during learning. It ensures the data's distribution, $\mathcal{N}\left(z^v;\mu^v,\left(\sigma^v\right)^2\right)$, resembles a standard bell curve, $\mathcal{N}(\epsilon ; 0, \mathbf{I})$. This similarity is gauged using Kullback-Leibler divergence (KLD) between the distributions, promoting model diversity and reducing uncertainty, leading to robust features. Eq~\ref{eq:reg} shows how $\mathcal{L}{v}$ is computed, with analogous methods for $\mathcal{L}{c}$ and $\mathcal{L}_{a}$, replacing $z^v$ with $z^c$ and $\hat{a}^k$.

\begin{equation}
\label{eq:reg}
\begin{aligned}
\mathcal{L}_v & =K L\left(\mathcal{N}\left(z^v ; \mu^v,\left(\sigma^v\right)^2\right) \| \mathcal{N}(\epsilon ; 0, \mathbf{I})\right) \\
& =-\frac{1}{2}\left(1+\log \left(\sigma^v\right)^2-\left(\mu^v\right)^2-\left(\sigma^v\right)^2\right)
\end{aligned}
\end{equation}

\section{Experimental Analysis}
\label{sec:exp}

\subsection{Datasets}




To compare with SAMUS~\cite{samus}, we test CC-SAM on the same seven public datasets: TN3K~\cite{tn3k}, DDTI~\cite{ddti}, TG3K~\cite{tg3k}, BUSI~\cite{busi}, UDIAT~\cite{udiat}, CAMUS~\cite{camus}, and HMC-QU~\cite{hmcqu}. We also recreate the US30K dataset from SAMUS for broader comparison. Dataset details are in the supplementary material.

TN3K and TG3K datasets follow TRFE~\cite{tn3k} for segmentation into training, validation, and testing. BUSI is randomly split in a 7:1:2 ratio for these purposes. CAMUS is initially divided into training and testing per the challenge~\cite{camus}, with 10\% of training data later used for validation.

For model generalizability, US30K datasets remain unseen during training and validation. In line with SAMUS, we split US30k into seen and unseen datasets. We then evaluate on unseen datasets (DDTI, UDIAT, HMC-QU) not used in training or validation. This enables direct comparison of generalization with other state-of-the-art models. Moreover, we contrast CC-SAM with other base models by training on the full US30K dataset and testing across different tasks.

\subsection{Implementation Details}

We follow SAMUS in setting hyper parameters and adapter sizes. We change the fully tuneable CNN in SAMUS to a static ResNet50. The model is trained using the Adam optimizer with an initial learning rate of 0.01 that reduces by a tenth every 50 epochs for a total of 200 epochs. The encoders are fully connected networks with the latent space having dimensions 1024. All fully connected layers has a Leaky RELU activation function and a dropout probability of 0.5.

We use 1xA100 GPU with 40 GB capacity. But we also verified that a single 3090 Ti with 24 GB memory is enough to train the model. We set a batch size of 32. 

\subsection{Ablation Study}
\label{sec:ablation}

We consider SAMUS~\cite{samus} as our baseline model and ablate our added components. Specifically, we show replacing the end-to-end CNN model with a fixed pre-trained model improves performance. Further adding the FC layer at the end slightly improves performance. Crucially, replacing the cross attention fusion with our variational attention fusion significantly improves performance. Further, replacing the point based prompting with our text-based prompting gives us the best results. We show these results on two of the datasets TN3K and BUSI in Table~\ref{tab:ablation}.

\begin{table}[htb]
\centering

\begin{tabular}{|ccccccc|}
\hline
\multicolumn{3}{|c|}{Components} & \multicolumn{2}{c|}{TN3K} & \multicolumn{2}{c|}{BUSI} \\ \hline
\multicolumn{1}{|c|}{ResNet50} & \multicolumn{1}{c|}{VAF} & \multicolumn{1}{c|}{Text Prompt} & \multicolumn{1}{c|}{Dice} & \multicolumn{1}{c|}{HD} & \multicolumn{1}{c|}{Dice} & \multicolumn{1}{c|}{HD} \\ \hline
$\times$ & $\times$ & $\times$ & 84.45 & 28.22 & 85.77 & 25.49 \\
\checkmark & $\times$ & $\times$ & 84.91 & 27.25 & 86.22 & 25.12 \\
$\times$ & \checkmark & $\times$ & 85.55 & 26.84 & 87.15 & 24.42 \\
$\times$ & $\times$ & \checkmark & 85.95 & 26.22 & 87.92 & 23.92 \\
\checkmark & $\times$ & \checkmark & 86.42 & 25.82 & 88.45 & 23.20 \\
$\times$ & \checkmark & \checkmark & 86.25 & 25.92 & 88.22 & 23.51 \\
\checkmark & \checkmark & $\times$ & 85.72 & 26.21 & 87.42 & 24.61 \\
\checkmark & \checkmark & \checkmark & \textbf{87.11} & \textbf{25.25} & \textbf{89.51} & \textbf{22.86} \\
\hline
\end{tabular}
\caption{Ablation study of adding different components of our proposed approach. The first row corresponds to the baseline SAMUS~\cite{samus}.}
\label{tab:ablation}
\end{table}

We do a futher analysis of how the prompting compares with different prompting strategies in the supplementary material.

\subsection{Comparison with state-of-the-art}

We consider two scenarios for comparing with state-of-the-art models following SAMUS~\cite{samus}. We compare CC-SAM with task-specific methods and foundational models.

\subsubsection{Task-specific methods}

Following SAMUS~\cite{samus}, thirteen state-of-the-art, task-specific techniques are chosen for evaluation, spanning CNN-based, transformer-based, and CNN-Transformer hybrid methods. Methods other than SAMUS include CNN-based techniques encompass U-Net~\cite{unet}, CPFNet~\cite{cpfnet}, CA-Net~\cite{canet}, CE-Net~\cite{cenet}, and AAU-Net~\cite{aaunet}. The transformer-centric methods consist of SwinUnet~\cite{swinunet}, SETR~\cite{setr}, and MISSFormer~\cite{missformer}. The hybrids of CNN and Transformer are TransUNet~\cite{transunet}, TransFuse~\cite{transfuse}, FAT-Net~\cite{fatnet}, and H2Former~\cite{h2former}.

\begin{table*}[ht]
\centering
\begin{tabular}{|l|c|c|c|c|c|c|c|c|c|c|}
\hline
Method & \multicolumn{2}{c|}{CAMUS-LA} & \multicolumn{2}{c|}{TN3K} & \multicolumn{2}{c|}{BUSI} & \multicolumn{2}{c|}{CAMUS-LV} & \multicolumn{2}{c|}{CAMUS-MYO} \\
\hline
 & Dice & HD & Dice & HD & Dice & HD & Dice & HD & Dice & HD \\
\hline
U-Net & 91.00 & 12.91 & 79.01 & 34.12 & 78.11 & 33.60 & 93.56 & 9.90 & 86.86 & 16.87 \\
CPFNet & 91.51 & 12.26 & 79.43 & 33.07 & 80.56 & 27.98 & 93.32 & 9.63 & 86.68 & 16.51 \\
CA-Net & 91.28 & 12.32 & 80.52 & 33.65 & 81.88 & 28.67 & 93.59 & 9.77 & 87.21 & 16.24 \\
CE-Net & 91.14 & 12.29 & 80.37 & 32.79 & 80.21 & 30.19 & 93.36 & 9.91 & 86.47 & 16.66 \\
AAU-Net & 91.33 & 12.12 & 82.28 & 30.53 & 80.81 & 30.39 & 93.32 & 9.97 & 86.98 & 16.49 \\
SwinUNet & 89.80 & 14.74 & 70.08 & 44.13 & 67.23 & 47.02 & 91.72 & 12.80 & 84.46 & 20.25 \\
SETR & 90.52 & 13.91 & 67.80 & 44.11 & 68.22 & 40.37 & 92.82 & 11.34 & 86.20 & 18.27 \\
MISSFormer & 91.18 & 11.82 & 79.42 & 32.85 & 78.43 & 33.10 & 93.25 & 9.94 & 86.57 & 16.50 \\
TransUNet & 91.37 & 12.46 & 81.44 & 30.98 & 82.22 & 27.54 & 93.60 & 9.60 & 87.20 & 17.25 \\
FAT-Net & 91.55 & 12.05 & 80.45 & 32.77 & 82.16 & 28.55 & 93.59 & 9.20 & 87.19 & \textbf{15.93} \\
H2Former & 90.98 & 11.92 & 82.48 & 30.58 & 81.48 & 27.84 & 93.44 & 9.60 & 87.31 & 16.60 \\
SAMUS & 91.58 & 11.60 & 84.45 & 28.22 & 85.77 & 25.49 & 93.73 & 9.79 & 87.46 & 16.74 \\
\textbf{CC-SAM} & \textbf{92.03} & \textbf{11.11} & \textbf{85.20} & \textbf{27.10} & \textbf{87.01} & \textbf{24.22} & \textbf{93.95} & \textbf{9.11} & \textbf{88.25} & 16.11 \\ 
\hline
\end{tabular}
\caption{Quantitative comparison between our CC-SAM method and the state-of-the-art (SOTA) task-specific techniques for segmenting thyroid nodules (TN3K), breast cancer (BUSI), left ventricle (CAMUS-LV), myocardium (CAMUS-MYO), and left atrium (CAMUS-LA). We assessed the performance using the Dice score (\%) and the Hausdorff distance (HD). The top-performing results are highlighted in bold.}
\label{tab:quant1}
\end{table*}

\begin{figure*}[!htb]
    \centering
    \includegraphics[width=\textwidth]{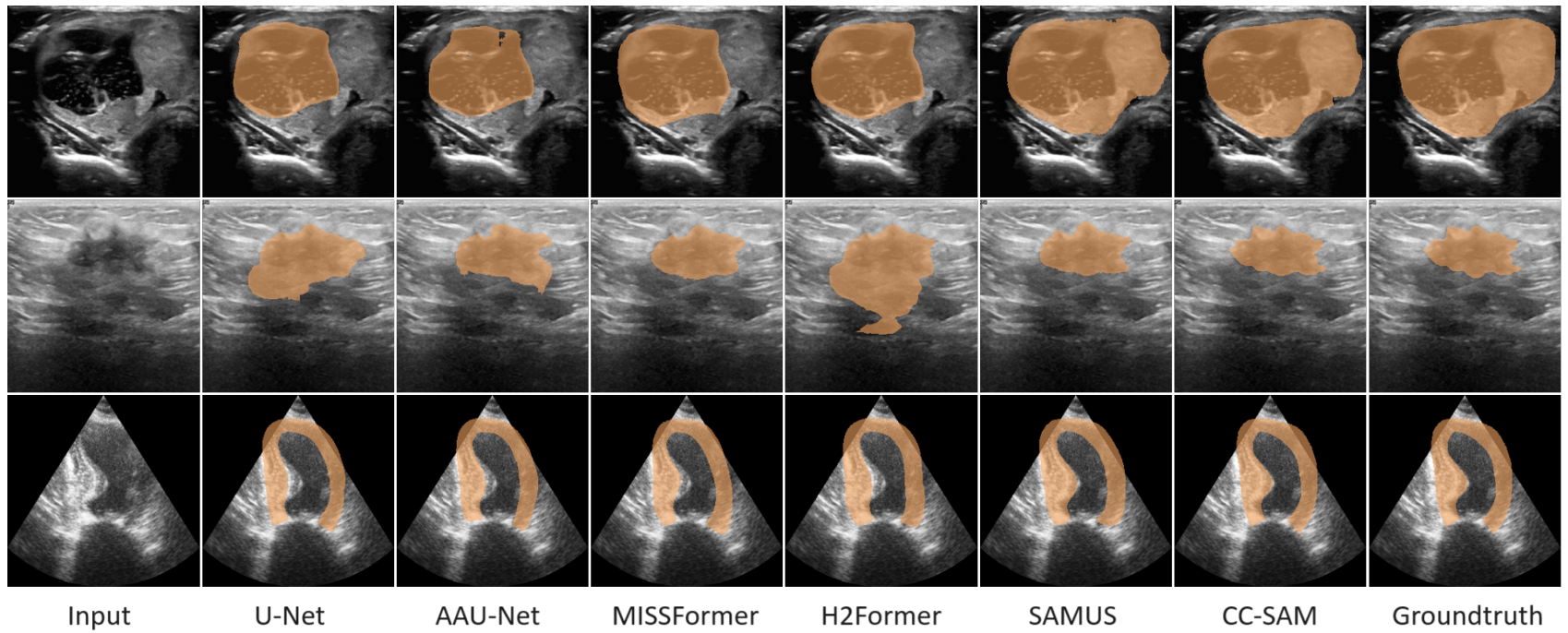}
    \caption{Qualitative comparison between our CC-SAM method and the state-of-the-art (SOTA) task-specific techniques.}
    \label{fig:qual1}
\end{figure*}

\paragraph{Quantitative Analysis.}

Table~\ref{tab:quant1} summarizes the numerical outcomes of various task-specific methods on TN3K, BUSI, CAMUS-LV, CAMUS-MYO, and CAMUS-LA. In comparison, CC-SAM consistently excels in performance across all these five tasks, setting a new benchmark for SOTA results. While CC-SAM surpasses recent SOTA in most benchmarks, it does fall short by a very small margin when evaluated with the Hausdorff distance (HD) on CAMUS-MYO.

\paragraph{Qualitative Analysis.}

Figure~\ref{fig:qual1} showcases the segmentation results from various methods, including U-Net~\cite{unet}, AAU-Net~\cite{aaunet}, MISSFormer~\cite{missformer}, H2Former~\cite{h2former}, SAMUS~\cite{samus}, and our proposed CC-SAM. Segmenting ultrasound images is notably challenging given their low contrast, inconsistent features, and indistinct object boundaries. Many existing techniques face difficulties in precisely differentiating the target from its background, often resulting in significant false negatives or false positives. In contrast, SAMUS stands out in maintaining the integrity of target areas and minimizing false positives. Our CC-SAM further refines the outcomes observed with SAMUS, a testament to the intrinsic strengths of SAM and the tailored modifications and innovations integrated into CC-SAM.

\paragraph{Generalization Ability.}

\begin{figure*}
    \centering
    \includegraphics[width=\textwidth]{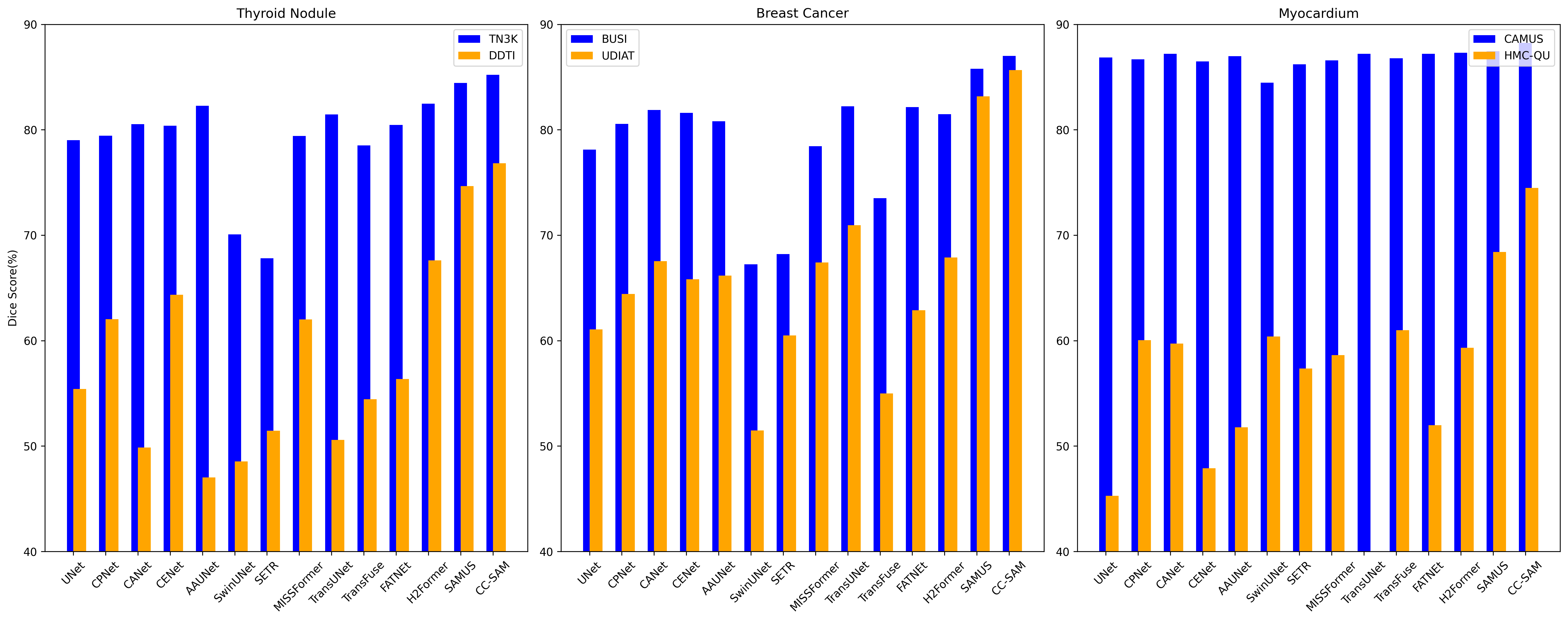}
    \caption{Comparison of CC-SAM with task-specific techniques on see datasets (highlighted in blue) and unseen datasets not previously encountered (indicated in orange). Higher orange bars indicate stronger generalization ability.}
    \label{fig:gen}
\end{figure*}

Evaluating the generalization performance of various task-specific methods is crucial as it provides insights into how well these methods can adapt and function across unfamiliar or unseen datasets. Figure~\ref{fig:gen} offers a quantitative assessment of this aspect. When we assess performance discrepancies between known (seeable) and unknown (unseen) datasets, CC-SAM stands out by displaying the most minimal performance drop among all methods across the three segmentation tasks. This underlines the remarkable generalization prowess of CC-SAM, emphasizing its resilience and versatility in a variety of medical image segmentation situations.

\subsubsection{Foundational models}

We also compare with five state-of-the-art foundational models. These include the original SAM~\cite{sam}, SAMed~\cite{samed}, SAMUS~\cite{samus}, MSA~\cite{msa} and MedSAM~\cite{medsam}. SAM is  trained on SA-1B and kept frozen due to the cost involved in fine-tuning. All other foundational models are pre-trained on the US30k dataset and evaluated using Dice and HD across TN3K, BUSI, CAMUS-LV, CAMUS-MYO,
and CAMUS-LA datasets.

\paragraph{Quantitative Analysis.}

\begin{table*}[ht]
\centering

\begin{tabular}{|c|c|c|c|c|c|c|c|c|c|c|}
\hline
Method & \multicolumn{2}{c|}{TN3K} & \multicolumn{2}{c|}{BUSI} & \multicolumn{2}{c|}{CAMUS-LV} & \multicolumn{2}{c|}{CAMUS-MYO} & \multicolumn{2}{c|}{CAMUS-LA} \\
\hline
 & Dice & HD & Dice & HD & Dice & HD & Dice & HD & Dice & HD \\
\hline
SAM & 29.59 & 134.87 & 54.01 & 82.39 & 28.18 & 196.64 & 29.42 & 184.10 & 17.28 & 193.70 \\
MedSAM & 71.09 & 42.91 & 77.75 & 34.26 & 87.52 & 15.28 & 76.07 & 25.72 & 88.06 & 15.70 \\
SAMed & 80.40 & 31.29 & 74.82 & 34.60 & 87.67 & 13.24 & 82.60 & 19.48 & 90.92 & 12.60 \\
MSA & 82.67 & 29.15 & 81.66 & 28.87 & 90.95 & 11.29 & 82.47 & 19.28 & 91.80 & 11.59 \\
SAMUS & 83.05 & 28.82 & 84.54 & 27.24 & 91.13 & 11.76 & 83.11 & 18.99 & 92.00 & 12.08 \\
\textbf{CC-SAM} & \textbf{85.59} & \textbf{27.74} & \textbf{86.22} & \textbf{25.85} & \textbf{92.85} & \textbf{10.88} & \textbf{85.61} & \textbf{17.11} & \textbf{93.51} & \textbf{11.06} \\
\hline
\end{tabular}

\caption{Quantitative comparison of our CC-SAM and other foundation models on seeable US30K data. The performance is evaluated by the Dice score (\%) and Hausdorff distance (HD).}
\label{tab:quant2}
\end{table*}

Table~\ref{tab:quant2} summarizes the numerical outcomes of various foundational methods on TN3K, BUSI, CAMUS-LV, CAMUS-MYO, and CAMUS-LA. In comparison, CC-SAM consistently excels in performance across all tasks, setting a new benchmark for foundational models.

\paragraph{Qualitative Analysis.}

\begin{figure*}
    \centering
    \includegraphics[width=\textwidth]{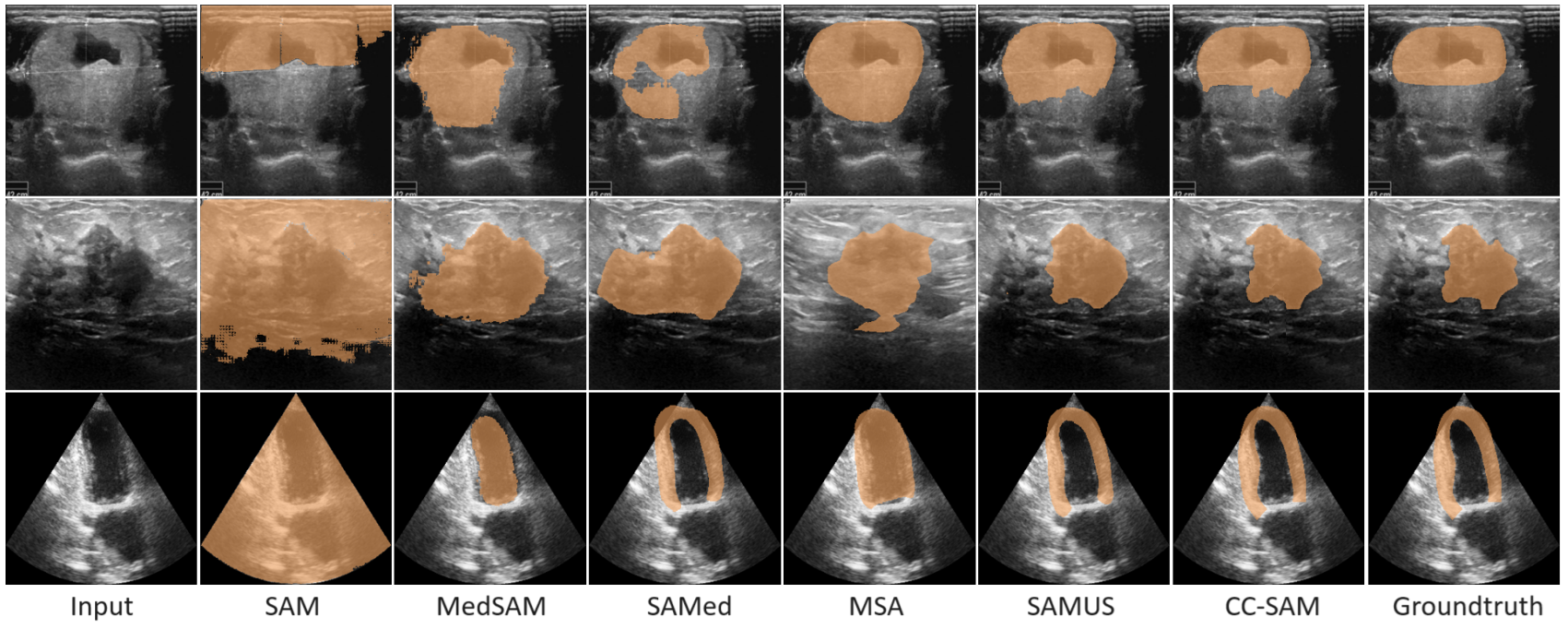}
    \caption{Qualitative comparison between our CC-SAM method and the state-of-the-art (SOTA) foundational medical segmentation models.}
    \label{fig:qual2}
\end{figure*}

Figure~\ref{fig:qual2} shows the qualitative segmentation results from various foundational models, including SAM, MedSAM, SAMed, MSA, SAMUS, and CC-SAM. SAM, without adjustments for medical images, fails at segmentation. However, MedSAM, SAMed, and MSA, through tuning, recover some segmentation ability but struggle with accurately defining borders, resulting in many false positives and negatives. In contrast, SAMUS excels in identifying precise segmentation borders, even in low contrast, suggesting the benefit of integrating local data with the image encoder for better boundary and shape retention in medical image segmentation. CC-SAM goes further, improving upon SAMUS's performance and setting new benchmarks.

\paragraph{Generalization Ability.}

\begin{figure}[h!]
    \centering
    \begin{minipage}{0.47\columnwidth}
        \centering
        \includegraphics[width=\linewidth]{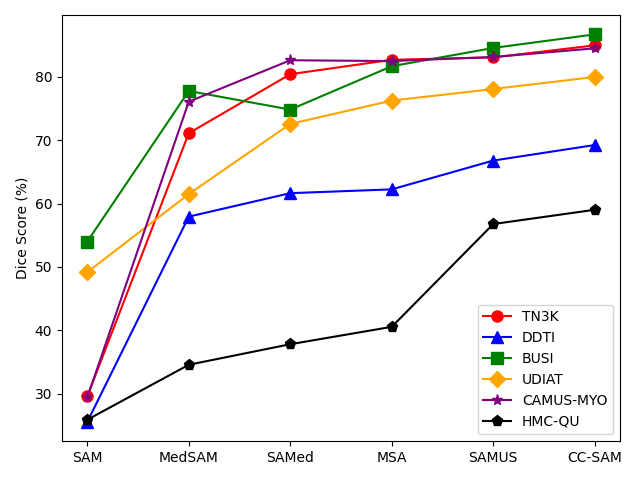}
        \caption{Comparing segmentation and generalization ability comparison of our CC-SAMUS and other foundation models. Whilst SAM obtains the lowest Dice Score, CC-SAM obtains the highest across all datasets.}
        \label{fig:gen2}
    \end{minipage}\hfill
    \begin{minipage}{0.49\columnwidth}
        \centering
        \includegraphics[width=\linewidth]{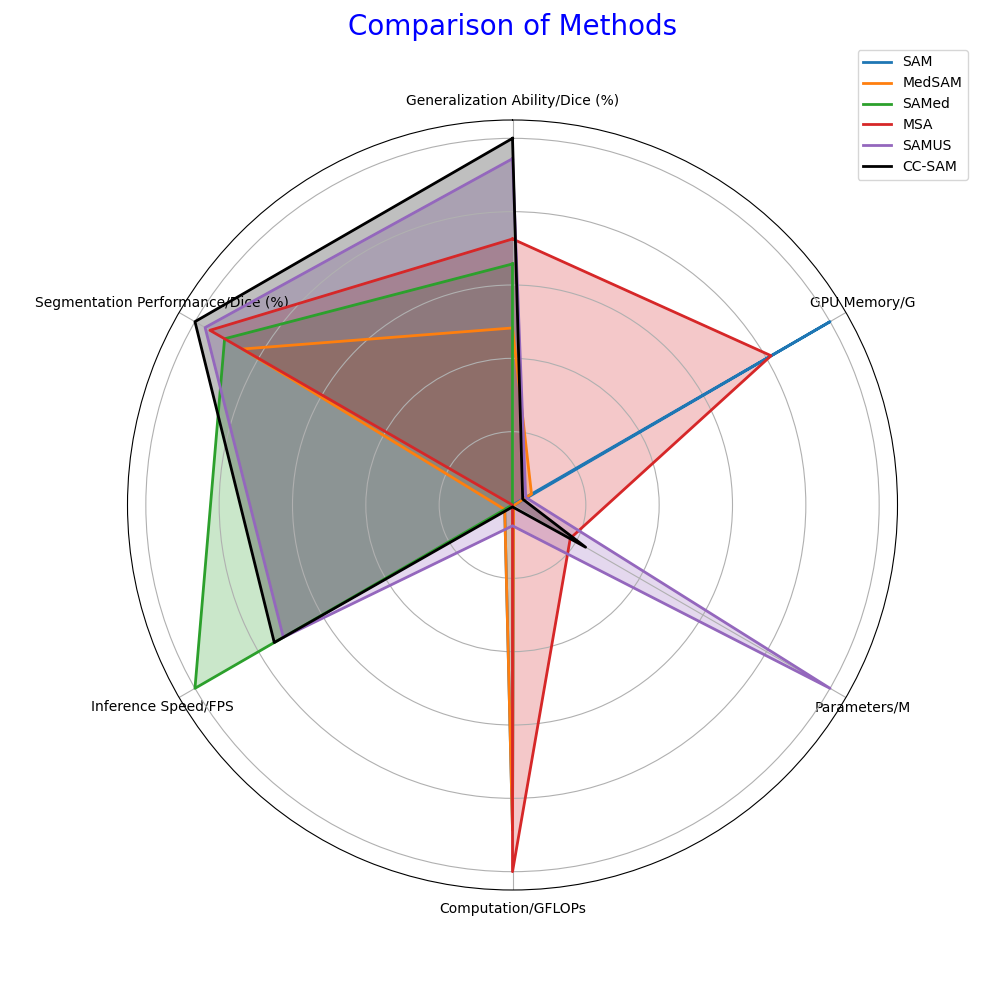}
        \caption{Spider plot comparing efficiency and segmentation performance of all foundational models.}
        \label{fig:spider}
    \end{minipage}
\end{figure}

Generally, foundation models trained on US30K exhibit a much improved generalization performance in medical image segmentation tasks compared to the original SAM. When considering the three specific segmentation tasks: thyroid nodule segmentation, breast cancer segmentation, and myocardium segmentation, it's clear that all foundation models face significant challenges with myocardium segmentation, while they perform relatively well in the breast cancer segmentation task. CC-SAM outperforms all foundational models significantly. This highlights the remarkable generalization prowess of CC-SAM, as it consistently surpasses other foundation models, especially in unfamiliar domains. Figure~\ref{fig:gen2} offers a quantitative assessment of this aspect. We see that the original SAM performs the worst, which is expected and the proposed CC-SAM obtains a new state-of-the-art on all datasets.

\paragraph{Deployment Cost Comparison with Other Foundational Models}

We compare CC-SAM with other foundational models in a more detailed manner. We look at `Generalization Ability/Dice (\%)', `GPU Memory/G', `Parameters/M', `Computation/GFLOPs', `Inference Speed/FPS' and `Segmentation Performance/Dice (\%)'.The spider plot of this can be seen in Figure~\ref{fig:spider}.

\section{Conclusion}

The Segment Anything Model (SAM), while proficient in natural image segmentation, faced hurdles in medical imaging due to issues with low contrast, faint boundaries, and other complexities. 
In this study, we identified and addressed the limitations of the Segment Anything Model (SAM) when applied to medical imaging, especially with images that have inherent challenges like low contrast and intricate morphologies. Our enhanced approach integrates a CNN branch with SAM's original Vision Transformer (ViT) encoder, utilizing a unique variational attention fusion module to capture vital local spatial information present in medical images. By introducing feature and position adapters within the ViT branch, the model's representations are further refined. Notably, we found that using text prompts, especially those generated leveraging ChatGPT's capabilities, significantly boosts SAM's performance in ultrasound medical segmentation. These innovations mark a pivotal step towards improving the adaptability and efficiency of universal image segmentation models in the medical field.

\paragraph{\textbf{Acknowledgements}}

DAC was supported by the Pandemic Sciences Institute at the University of Oxford; the National Institute for Health Research (NIHR) Oxford Biomedical Research Centre (BRC); an NIHR Research Professorship; a Royal Academy of Engineering Research Chair; the Wellcome Trust funded VITAL project (grant 204904/Z/16/Z); the EPSRC (grant EP/W031744/1); and the InnoHK Hong Kong Centre for Cerebro-cardiovascular Engineering (COCHE).

%
%
\bibliographystyle{splncs04}
\bibliography{main}

\begin{thebibliography}{10}
\providecommand{\url}[1]{\texttt{#1}}
\providecommand{\urlprefix}{URL }
\providecommand{\doi}[1]{https://doi.org/#1}

\bibitem{busi}
Al-Dhabyani, W., Gomaa, M., Khaled, H., Aly, F.: Deep learning approaches for data augmentation and classification of breast masses using ultrasound images. Int. J. Adv. Comput. Sci. Appl  \textbf{10}(5),  1--11 (2019)

\bibitem{vib}
Alemi, A.A., Fischer, I., Dillon, J.V., Murphy, K.: Deep variational information bottleneck. arXiv preprint arXiv:1612.00410  (2016)

\bibitem{udiat}
Byra, M., Jarosik, P., Szubert, A., Galperin, M., Ojeda-Fournier, H., Olson, L., O’Boyle, M., Comstock, C., Andre, M.: Breast mass segmentation in ultrasound with selective kernel u-net convolutional neural network. Biomedical Signal Processing and Control  \textbf{61},  102027 (2020)

\bibitem{swinunet}
Cao, H., Wang, Y., Chen, J., Jiang, D., Zhang, X., Tian, Q., Wang, M.: Swin-unet: Unet-like pure transformer for medical image segmentation. In: European conference on computer vision. pp. 205--218. Springer (2022)

\bibitem{aaunet}
Chen, G., Li, L., Dai, Y., Zhang, J., Yap, M.H.: Aau-net: an adaptive attention u-net for breast lesions segmentation in ultrasound images. IEEE Transactions on Medical Imaging  (2022)

\bibitem{transunet}
Chen, J., Lu, Y., Yu, Q., Luo, X., Adeli, E., Wang, Y., Lu, L., Yuille, A.L., Zhou, Y.: Transunet: Transformers make strong encoders for medical image segmentation. arXiv preprint arXiv:2102.04306  (2021)

\bibitem{chen2018drinet}
Chen, L., Bentley, P., Mori, K., Misawa, K., Fujiwara, M., Rueckert, D.: Drinet for medical image segmentation. IEEE transactions on medical imaging  \textbf{37}(11),  2453--2462 (2018)

\bibitem{hmcqu}
Degerli, A., Kiranyaz, S., Hamid, T., Mazhar, R., Gabbouj, M.: Early myocardial infarction detection over multi-view echocardiography. Biomedical Signal Processing and Control  \textbf{87},  105448 (2024). \doi{https://doi.org/10.1016/j.bspc.2023.105448}

\bibitem{vit}
Dosovitskiy, A., Beyer, L., Kolesnikov, A., Weissenborn, D., Zhai, X., Unterthiner, T., Dehghani, M., Minderer, M., Heigold, G., Gelly, S., et~al.: An image is worth 16x16 words: Transformers for image recognition at scale. arXiv preprint arXiv:2010.11929  (2020)

\bibitem{adapter5}
Ermis, B., Zappella, G., Wistuba, M., Rawal, A., Archambeau, C.: Continual learning with transformers for image classification. In: Proceedings of the IEEE/CVF Conference on Computer Vision and Pattern Recognition. pp. 3774--3781 (2022)

\bibitem{cpfnet}
Feng, S., Zhao, H., Shi, F., Cheng, X., Wang, M., Ma, Y., Xiang, D., Zhu, W., Chen, X.: Cpfnet: Context pyramid fusion network for medical image segmentation. IEEE transactions on medical imaging  \textbf{39}(10),  3008--3018 (2020)

\bibitem{adapter6}
Gao, P., Geng, S., Zhang, R., Ma, T., Fang, R., Zhang, Y., Li, H., Qiao, Y.: Clip-adapter: Better vision-language models with feature adapters. International Journal of Computer Vision pp. 1--15 (2023)

\bibitem{adapter4}
Gao, P., Han, J., Zhang, R., Lin, Z., Geng, S., Zhou, A., Zhang, W., Lu, P., He, C., Yue, X., et~al.: Llama-adapter v2: Parameter-efficient visual instruction model. arXiv preprint arXiv:2304.15010  (2023)

\bibitem{tn3k}
Gong, H., Chen, J., Chen, G., Li, H., Li, G., Chen, F.: Thyroid region prior guided attention for ultrasound segmentation of thyroid nodules. Computers in Biology and Medicine  \textbf{155},  106389 (2023)

\bibitem{gowda2017human}
Gowda, S.N.: Human activity recognition using combinatorial deep belief networks. In: Proceedings of the IEEE conference on computer vision and pattern recognition workshops. pp.~1--6 (2017)

\bibitem{adapter8}
Gowda, S.N., Arnab, A., Huang, J.: Optimizing vivit training: Time and memory reduction for action recognition. arXiv preprint arXiv:2306.04822  (2023)

\bibitem{gowda2024masks}
Gowda, S.N., Clifton, D.A.: Masks and manuscripts: Advancing medical pre-training with end-to-end masking and narrative structuring. arXiv preprint arXiv:2407.16264  (2024)

\bibitem{adapter9}
Gowda, S.N., Gao, B., Clifton, D.A.: Fe-adapter: Adapting image-based emotion classifiers to videos. In: 2024 IEEE 18th International Conference on Automatic Face and Gesture Recognition (FG). pp.~1--6. IEEE (2024)

\bibitem{gowda2021smart}
Gowda, S.N., Rohrbach, M., Sevilla-Lara, L.: Smart frame selection for action recognition. In: Proceedings of the AAAI Conference on Artificial Intelligence. vol.~35, pp. 1451--1459 (2021)

\bibitem{colornet}
Gowda, S.N., Yuan, C.: Colornet: Investigating the importance of color spaces for image classification. In: Computer Vision--ACCV 2018: 14th Asian Conference on Computer Vision, Perth, Australia, December 2--6, 2018, Revised Selected Papers, Part IV 14. pp. 581--596. Springer (2019)

\bibitem{canet}
Gu, R., Wang, G., Song, T., Huang, R., Aertsen, M., Deprest, J., Ourselin, S., Vercauteren, T., Zhang, S.: Ca-net: Comprehensive attention convolutional neural networks for explainable medical image segmentation. IEEE transactions on medical imaging  \textbf{40}(2),  699--711 (2020)

\bibitem{cenet}
Gu, Z., Cheng, J., Fu, H., Zhou, K., Hao, H., Zhao, Y., Zhang, T., Gao, S., Liu, J.: Ce-net: Context encoder network for 2d medical image segmentation. IEEE transactions on medical imaging  \textbf{38}(10),  2281--2292 (2019)

\bibitem{unetr}
Hatamizadeh, A., Tang, Y., Nath, V., Yang, D., Myronenko, A., Landman, B., Roth, H.R., Xu, D.: Unetr: Transformers for 3d medical image segmentation. In: Proceedings of the IEEE/CVF winter conference on applications of computer vision. pp. 574--584 (2022)

\bibitem{h2former}
He, A., Wang, K., Li, T., Du, C., Xia, S., Fu, H.: H2former: An efficient hierarchical hybrid transformer for medical image segmentation. IEEE Transactions on Medical Imaging  (2023)

\bibitem{resnet}
He, K., Zhang, X., Ren, S., Sun, J.: Deep residual learning for image recognition. In: Proceedings of the IEEE conference on computer vision and pattern recognition. pp. 770--778 (2016)

\bibitem{sammed2}
He, S., Bao, R., Li, J., Grant, P.E., Ou, Y.: Accuracy of segment-anything model (sam) in medical image segmentation tasks. arXiv preprint arXiv:2304.09324  (2023)

\bibitem{adapter}
Houlsby, N., Giurgiu, A., Jastrzebski, S., Morrone, B., De~Laroussilhe, Q., Gesmundo, A., Attariyan, M., Gelly, S.: Parameter-efficient transfer learning for nlp. In: International Conference on Machine Learning. pp. 2790--2799. PMLR (2019)

\bibitem{densenet}
Huang, G., Liu, Z., Van Der~Maaten, L., Weinberger, K.Q.: Densely connected convolutional networks. In: Proceedings of the IEEE conference on computer vision and pattern recognition. pp. 4700--4708 (2017)

\bibitem{missformer}
Huang, X., Deng, Z., Li, D., Yuan, X.: Missformer: An effective medical image segmentation transformer. arXiv preprint arXiv:2109.07162  (2021)

\bibitem{huang2023segment}
Huang, Y., Yang, X., Liu, L., Zhou, H., Chang, A., Zhou, X., Chen, R., Yu, J., Chen, J., Chen, C., et~al.: Segment anything model for medical images? arXiv preprint arXiv:2304.14660  (2023)

\bibitem{repar}
Kingma, D.P., Welling, M.: Auto-encoding variational bayes. arXiv preprint arXiv:1312.6114  (2013)

\bibitem{sam}
Kirillov, A., Mintun, E., Ravi, N., Mao, H., Rolland, C., Gustafson, L., Xiao, T., Whitehead, S., Berg, A.C., Lo, W.Y., et~al.: Segment anything. arXiv preprint arXiv:2304.02643  (2023)

\bibitem{camus}
Leclerc, S., Smistad, E., Pedrosa, J., {\O}stvik, A., Cervenansky, F., Espinosa, F., Espeland, T., Berg, E.A.R., Jodoin, P.M., Grenier, T., et~al.: Deep learning for segmentation using an open large-scale dataset in 2d echocardiography. IEEE transactions on medical imaging  \textbf{38}(9),  2198--2210 (2019)

\bibitem{samus}
Lin, X., Xiang, Y., Zhang, L., Yang, X., Yan, Z., Yu, L.: Samus: Adapting segment anything model for clinically-friendly and generalizable ultrasound image segmentation. arXiv preprint arXiv:2309.06824  (2023)

\bibitem{gdino}
Liu, S., Zeng, Z., Ren, T., Li, F., Zhang, H., Yang, J., Li, C., Yang, J., Su, H., Zhu, J., et~al.: Grounding dino: Marrying dino with grounded pre-training for open-set object detection. arXiv preprint arXiv:2303.05499  (2023)

\bibitem{liu2021review}
Liu, X., Song, L., Liu, S., Zhang, Y.: A review of deep-learning-based medical image segmentation methods. Sustainability  \textbf{13}(3), ~1224 (2021)

\bibitem{adapter3}
Luo, Z., Hu, Z., Xi, Y., Zhang, R., Ma, J.: I-tuning: Tuning frozen language models with image for lightweight image captioning. In: ICASSP 2023-2023 IEEE International Conference on Acoustics, Speech and Signal Processing (ICASSP). pp.~1--5. IEEE (2023)

\bibitem{medsam}
Ma, J., Wang, B.: Segment anything in medical images. arXiv preprint arXiv:2304.12306  (2023)

\bibitem{rin}
Mei, X., Liu, Z., Robson, P.M., Marinelli, B., Huang, M., Doshi, A., Jacobi, A., Cao, C., Link, K.E., Yang, T., et~al.: Radimagenet: an open radiologic deep learning research dataset for effective transfer learning. Radiology: Artificial Intelligence  \textbf{4}(5),  e210315 (2022)

\bibitem{gpt4}
OpenAI: Gpt-4 technical report (2023)

\bibitem{adapter7}
Pan, J., Lin, Z., Zhu, X., Shao, J., Li, H.: St-adapter: Parameter-efficient image-to-video transfer learning. Advances in Neural Information Processing Systems  \textbf{35},  26462--26477 (2022)

\bibitem{ddti}
Pedraza, L., Vargas, C., Narv{\'a}ez, F., Dur{\'a}n, O., Mu{\~n}oz, E., Romero, E.: An open access thyroid ultrasound image database. In: 10th International symposium on medical information processing and analysis. vol.~9287, pp. 188--193. SPIE (2015)

\bibitem{adapter1}
Pfeiffer, J., R{\"u}ckl{\'e}, A., Poth, C., Kamath, A., Vuli{\'c}, I., Ruder, S., Cho, K., Gurevych, I.: Adapterhub: A framework for adapting transformers. arXiv preprint arXiv:2007.07779  (2020)

\bibitem{pham2000current}
Pham, D.L., Xu, C., Prince, J.L.: Current methods in medical image segmentation. Annual review of biomedical engineering  \textbf{2}(1),  315--337 (2000)

\bibitem{medbert}
Rasmy, L., Xiang, Y., Xie, Z., Tao, C., Zhi, D.: Med-bert: pretrained contextualized embeddings on large-scale structured electronic health records for disease prediction. NPJ digital medicine  \textbf{4}(1), ~86 (2021)

\bibitem{unet}
Ronneberger, O., Fischer, P., Brox, T.: U-net: Convolutional networks for biomedical image segmentation. In: Medical Image Computing and Computer-Assisted Intervention--MICCAI 2015: 18th International Conference, Munich, Germany, October 5-9, 2015, Proceedings, Part III 18. pp. 234--241. Springer (2015)

\bibitem{sammed1}
Roy, S., Wald, T., Koehler, G., Rokuss, M.R., Disch, N., Holzschuh, J., Zimmerer, D., Maier-Hein, K.H.: Sam. md: Zero-shot medical image segmentation capabilities of the segment anything model. arXiv preprint arXiv:2304.05396  (2023)

\bibitem{adapter2}
Sung, Y.L., Cho, J., Bansal, M.: Vl-adapter: Parameter-efficient transfer learning for vision-and-language tasks. In: Proceedings of the IEEE/CVF Conference on Computer Vision and Pattern Recognition. pp. 5227--5237 (2022)

\bibitem{unext}
Valanarasu, J.M.J., Patel, V.M.: Unext: Mlp-based rapid medical image segmentation network. In: International Conference on Medical Image Computing and Computer-Assisted Intervention. pp. 23--33. Springer (2022)

\bibitem{wang2022adamix}
Wang, Y., Mukherjee, S., Liu, X., Gao, J., Awadallah, A.H., Gao, J.: Adamix: Mixture-of-adapter for parameter-efficient tuning of large language models. arXiv preprint arXiv:2205.12410  \textbf{1}(2), ~4 (2022)

\bibitem{fatnet}
Wu, H., Chen, S., Chen, G., Wang, W., Lei, B., Wen, Z.: Fat-net: Feature adaptive transformers for automated skin lesion segmentation. Medical image analysis  \textbf{76},  102327 (2022)

\bibitem{msa}
Wu, J., Fu, R., Fang, H., Liu, Y., Wang, Z., Xu, Y., Jin, Y., Arbel, T.: Medical sam adapter: Adapting segment anything model for medical image segmentation. arXiv preprint arXiv:2304.12620  (2023)

\bibitem{tg3k}
Wunderling, T., Golla, B., Poudel, P., Arens, C., Friebe, M., Hansen, C.: Comparison of thyroid segmentation techniques for 3d ultrasound. In: Medical Imaging 2017: Image Processing. vol. 10133, pp. 346--352. SPIE (2017)

\bibitem{samed}
Zhang, K., Liu, D.: Customized segment anything model for medical image segmentation. arXiv preprint arXiv:2304.13785  (2023)

\bibitem{transfuse}
Zhang, Y., Liu, H., Hu, Q.: Transfuse: Fusing transformers and cnns for medical image segmentation. In: Medical Image Computing and Computer Assisted Intervention--MICCAI 2021: 24th International Conference, Strasbourg, France, September 27--October 1, 2021, Proceedings, Part I 24. pp. 14--24. Springer (2021)

\bibitem{setr}
Zheng, S., Lu, J., Zhao, H., Zhu, X., Luo, Z., Wang, Y., Fu, Y., Feng, J., Xiang, T., Torr, P.H., et~al.: Rethinking semantic segmentation from a sequence-to-sequence perspective with transformers. In: Proceedings of the IEEE/CVF conference on computer vision and pattern recognition. pp. 6881--6890 (2021)

\bibitem{zhou2018unet++}
Zhou, Z., Rahman~Siddiquee, M.M., Tajbakhsh, N., Liang, J.: Unet++: A nested u-net architecture for medical image segmentation. In: Deep Learning in Medical Image Analysis and Multimodal Learning for Clinical Decision Support: 4th International Workshop, DLMIA 2018, and 8th International Workshop, ML-CDS 2018, Held in Conjunction with MICCAI 2018, Granada, Spain, September 20, 2018, Proceedings 4. pp. 3--11. Springer (2018)

\end{thebibliography}
\end{document}